\documentclass{article}
\usepackage{arxiv}

\usepackage[utf8]{inputenc} 
\usepackage[T1]{fontenc}    
\usepackage{hyperref}       
\usepackage{url}            
\usepackage{booktabs}       
\usepackage{amsfonts}       
\usepackage{nicefrac}       
\usepackage{microtype}      
\usepackage{lipsum}

\usepackage{times}
\usepackage{natbib}
\setcitestyle{authoryear,open={(},close={)}}

\newcommand{\mbert}{mBERT}
\newcommand{\bbert}{bBERT}
\newcommand{\finbert}{FinBERT}
\newcommand{\bertbase}{BERT-Base}
\newcommand{\idt}{\hspace{1em}}

\usepackage{todonotes}

\title{Towards Fully Bilingual Deep Language Modeling}

\author{
  Li-Hsin~Chang \\
  \texttt{lhchan@utu.fi} \\
   \And
 Sampo~Pyysalo \\
  \texttt{sampyy@utu.fi} \\
   \And
 Jenna~Kanerva \\
  \texttt{jmnybl@utu.fi} \\
   \And
 Filip~Ginter \\
  \texttt{figint@utu.fi} \\
   \And
   \\
  TurkuNLP Group \\
  Department of Future Technologies\\
  University of Turku\\
  Turku, Finland \\
}
  
\date{}

\begin{document}
\maketitle
\begin{abstract}
Language models based on deep neural networks have facilitated great advances in natural language processing and understanding tasks in recent years. While models covering a large number of languages have been introduced, their multilinguality has come at a cost in terms of monolingual performance, and the best-performing models at most tasks not involving cross-lingual transfer remain monolingual. In this paper, we consider the question of whether it is possible to pre-train a bilingual model for two remotely related languages without compromising performance at either language. We collect pre-training data, create a Finnish-English bilingual BERT model and evaluate its performance on datasets used to evaluate the corresponding monolingual models. Our bilingual model performs on par with Google's original English BERT on GLUE and nearly matches the performance of monolingual Finnish BERT on a range of Finnish NLP tasks, clearly outperforming multilingual BERT.
We find that when the model vocabulary size is increased, the \bertbase{} architecture has sufficient capacity to learn two remotely related languages to a level where it achieves comparable performance with monolingual models, demonstrating the feasibility of training fully bilingual deep language models.
The model and all tools involved in its creation are freely available at \url{https://github.com/TurkuNLP/biBERT}
\end{abstract}


\section{Introduction}
\label{sec:introduction}

In recent years, there has been an increased focus on the use of unannotated texts for modeling human language and on transfer learning in natural language processing (NLP). A wide variety of models have been proposed, ranging from context-independent word embeddings \citep{Mikolov2013w2v,pennington2014glove}, to the more recent contextual representations \citep{peters2018deep,devlin2018bert}. In particular, the Transformer-based \citep{vaswani2017attention} BERT (Bidirectional Encoder Representations from Transformers) model \citep{devlin2018bert} has generated considerable interest in the NLP community since its release. BERT outperformed the then state-of-the-art systems on a wide range of benchmark datasets when published, and has served as the basis of many studies since. These efforts include work that proposes improvements and/or modifications to the training objectives \citep{Liu2019roberta,Lan2020alberta}, knowledge distillation \citep{Sanh2019DistilBERT}, multilinguality \citep{Pires2019HowMBERT}, and interpretation \citep{Kovaleva2019BERTdarkSecrets}, to name a few. As a mark of its popularity, the term BERTology was coined to refer to the field of research relating to BERT \citep{Rogers2020BERTology}.

A thriving branch of BERTology involves BERT models for languages other than English. \cite{devlin2018bert} released multilingual BERT (\mbert{}) models trained on over a hundred languages.
\cite{Wu2019xlingBERT} analyze the representations produced by mulitlingual BERTs and find evidence that these representations generalize across languages for various downstream tasks, though language-specific information is retained. This language-agnostic subspace of multingual BERTs has also been observed in other studies, and is deemed to be the factor that allows for zero-shot transfer \citep{Pires2019HowMBERT,Cao2020mBERTAlignment}.
Furthermore, the embeddings can be further aligned through a fine-tuning based alignment procedure, improving the performance of multilingual models \citep{Cao2020mBERTAlignment}.
While multilingual training can benefit also monolingual performance, as the number of languages covered by a multilingual model increases, the fraction of the model capacity available for any single language decreases. 
\cite{Conneau2019XLMR} term as the \emph{curse of multilinguality} the phenomenon where increasing the number of languages included in a model initially leads to better cross-lingual performance for low-resource languages, while eventually leading to overall degradation of both monolingual and cross-lingual performance.
Work on language-specific BERT models has also shown that monolingual models tend to outperform multilingual models of the same size in monolingual settings \citep{de2019bertje,martin2020camembert,virtanen2019multilingual,Pyysalo2020WikiBERT}.
However, the question of whether it is possible to train multilingual models without loss of monolingual performance remains largely open.

In this paper, we study whether it is feasible to pre-train a bilingual model for two remotely related languages without compromising performance at either language. Specifically, we train a Finnish-English bilingual BERT model (henceforth, \bbert{}) using a combination of the pre-training data of the original English BERT model and the Finnish BERT model introduced by \cite{virtanen2019multilingual}, using an extended model vocabulary but otherwise fixing model capacity at BERT-Base size and retaining the number of pre-training steps.
We evaluate the performance of the introduced bilingual model on a range of natural language understanding (NLU) tasks used to evaluate the monolingual models, which, to the best of our knowledge, has not been the focus of studies on bilingual BERT models. We find that \bbert{} achieves comparable performance on the GLUE (General Language Understanding Evaluation) benchmark \citep{wang2018glue} with the original English BERT, and nearly matches the performance of the Finnish BERT on Finnish NLP tasks. Our results indicate that an extension of the vocabulary size is sufficient to allow the creation of fully bilingual models that perform on par with their monolingual counterparts in both of their languages.

\section{Related work} 
\label{sec:related-work}
The BERT variants available can be categorized according to the number of languages they are trained on: monolingual BERTs, multilingual BERTs with few languages, and multilingual BERTs with a large number of languages. The original authors of BERT released several versions of BERT varying in the model size, casing, and language \citep{devlin2018bert}. Among these models is the cased English \bertbase{} model. 
Its architecture, the \bertbase{} model architecture, has 12 layers with hidden dimension of 768 and 12 attention heads, resulting in a total of 110M parameters.
The English BERT was evaluated on GLUE, the Stanford Question Answering Dataset (SQuAD) \citep{rajpurkar2016squad}, and the Situations With Adversarial Generations dataset \citep{zellers2018SWAG}.

Other monolingual BERT models have been trained and released by the NLP community. \cite{virtanen2019multilingual} train Finnish cased and uncased versions of \bertbase{} models (the cased Finnish model is referred to as \finbert{} henceforth). The Finnish BERTs have been evaluated on part-of-speech (POS) tagging, named entity recognition (NER), dependency parsing, text classification, and probing tasks. \finbert{} outperforms \mbert{} on nearly all of these tasks, illustrating the advantages of monolingual models over multilingual ones. To benefit a wider range of languages, \cite{Pyysalo2020WikiBERT} construct an automatic pipeline to train monolingual BERTs on Wikipedia. They train 42 monolingual BERT models using this pipeline and tested their parsing performance. They find that while language-specific models lead to improvement in parsing performance on average, the relative performance of the models vary substantially depending on the language.

Apart from monolingual models that have been trained for languages with more resources, the cross-lingual transferability of models has also been studied to empower languages with fewer resources. \cite{Artetxe2020cross} successfully transfer monolingual representations to other languages by freezing the model weights and retraining only the vocabulary weights. However, the performance of these cross-lingually transferred monolingual models tend not to match that of bilingual models in their experiments. Thus, for languages with sufficient resources to train monolingual BERTs that do not have as much labelled data as English, training bilingual BERTs to combine the benefit of BERT's language-agnostic subspace and the advantage of models with fewer languages to avoid the curse of multilinguality presents a potential solution. Whether the monolingual capability of such models is compromised, however, remains an open question.

The studies on multilingual BERT models with few languages have mainly focused on cross-lingual aspects. \cite{Karthikeyan2020BBERT} study the cross-lingual capability of multilingual BERTs by training bilingual BERT models and studying the effect of linguistic properties of languages, model architectures, and learning objectives. Their models are evaluated with NER and textual entailment tasks. Compared to our study, their focus is on cross-linguality, and their models are trained for a lesser number of steps than ours. In work performed concurrently with ours, \cite{Ulcar2020FinEstBERT} train Finnish-Estonian-English and Croatian-Slovenian-English trilingual BERT models and evaluate their performance on POS tagging, NER, and dependency parsing. Their baselines are multilingual contextual representation models; there is no evaluation of monolingual model performance. This focus of evaluation on the cross-lingual ability of the models differs from our focus, which is placed on the language-specific capability of bilingual models compared to monolingual models. 

There have also been studies on multilingual BERTs with a larger number of languages, though these models tend to suffer from the curse of multilinguality. In the study of \cite{Artetxe2020cross}, bilingual BERT models tend to outperform their multilingual BERT model jointly trained on 15 languages. \cite{devlin2018bert} release cased and uncased multilingual \bertbase{} models trained on over a hundred languages (the multilingual cased model \mbert{} henceforth). These multilingual models, however, tend to underperform monolingual models \citep{de2019bertje,martin2020camembert,virtanen2019multilingual,Pyysalo2020WikiBERT}.

\section{Pre-training}
\label{sec:pretraining}
This section introduces the sources of unannotated English and Finnish texts used for pre-training, as well as the collection and filtering of these texts, the generation of the model vocabulary, and the pre-training process.

\subsection{Pre-training data}

The original English BERT was trained on the BooksCorpus \citep{Zhu2015BC} and English Wikipedia, which consist of 800M words and 2,500M words respectively. Since BooksCorpus is no longer available, we reconstruct an approximation from URLs collected in a separate crawl of the corpus sources.\footnote{\url{https://github.com/soskek/bookcorpus}}
The collected books are filtered to exclude non-English text by language detection. Hand-written heuristics are used to remove short sentences and sentences with high ratios of uppercase characters, digits, or foreign characters. Since the data came from books, tables of contents, copyright messages, and references are likewise removed by hand-written heuristics. Finally, content duplications are removed with the corpus tool Onion \citep{pomikalek2011onion}.
The English Wikipedia data were obtained using parts of the pipeline introduced by \cite{Pyysalo2020WikiBERT}.\footnote{\url{https://github.com/spyysalo/wiki-bert-pipeline}} More specifically, these components were used to download a Wikipedia database backup dump, extract plain text from the XML sources using WikiExtractor\footnote{\url{https://github.com/attardi/wikiextractor}}, segment and tokenize the text, and perform heuristic document filtering.

For the Finnish pre-training data, we use the same data used for training FinBERT \citep{virtanen2019multilingual}. Briefly, the data come from three sources: news from Yle and the Finnish News Agency, online discussion from the Suomi24 forum, and an internet crawl. After filtering and deduplication, the data is about 30 times the size that of the Finnish Wikipedia included in the data that \mbert{} was trained on. A Detailed description of the data and its preprocessing can be found in \cite{virtanen2019multilingual}.

\begin{table}[!t]
\centering
\begin{tabular}{lrrr}
\toprule
Data & Sentences & Tokens \\
\midrule
English & 198M & 3.8B \\
\idt Wikipedia & 130M & 2.8B \\
\idt BooksCorpus (reconstructed) & 68M & 1.0B \\ 
Finnish & 234M & 3.3B \\
\idt News & 36M & 0.5B \\
\idt Online discussion & 118M & 1.7B \\
\idt Internet crawl & 79M & 1.1B \\
\bottomrule
\end{tabular}
\caption{Statistics for pre-training data}
\label{tab:pretraining_stats}
\end{table}

The statistics of the pre-training data are presented in Table~\ref{tab:pretraining_stats}. We note that in terms of the total number of tokens, the data sources for the two languages are remarkably closely balanced, with 3.8B tokens for English and 3.3B for Finnish.

\subsection{Vocabulary generation}

For vocabulary generation, we take a sample of the cleaned and filtered sentences and tokenize them using BERT BasicTokenizer. To balance the vocabulary for the two languages, the same number of sentences are sampled for each language, proportionally to the size of the source. In total, 10 million sentences are sampled, half of which are English and the other half Finnish. 
Due to downstream evaluation results reported by \cite{virtanen2019multilingual} and others tending to favour cased over uncased models, we here chose to train a cased bilingual model. The Sentence-Piece \citep{Kudo2018SentencePiece} implementation of byte-pair-encoding \citep{Sennrich2016BPE} is used to generate the bilingual vocabulary. The generated vocabulary is then converted to a WordPiece \citep{Wu2016WordPiece} vocabulary.
Taking into account the observation of \cite{Artetxe2020cross} that the effective vocabulary size per language plays a more important role in the model performance than either the choice between a joint or disjoint vocabulary or the number of languages for multilingual models, we fix the bilingual vocabulary to be a joint vocabulary of 80,000 words, matching the combined size of the English BERT (30,000 words) and FinBERT (50,000 words) vocabularies.

Comparing the bilingual vocabulary to the monolingual ones, we find that the \bbert{} vocabulary contains 87.5\% of the WordPieces in the original Google BERT vocabulary, and 61.5\% of the WordPieces in the FinBERT vocabulary. The lower coverage of FinBERT vocabulary is expected as the sampling strategy for vocabulary generation balances English and Finnish, whereas the English BERT vocabulary is smaller than that of the Finnish BERT.

\subsection{Pre-training example generation}

Following the approaches of \cite{devlin2018bert} and \cite{virtanen2019multilingual}, the pre-training examples are created for the masked language modeling and next sentence prediction tasks. Duplication factors are set so that there are roughly the same number of training examples for Finnish and English, and that the number of examples covered in the whole training data matches that of FinBERT. For Finnish, each source (news, discussion, and crawl) has a separate duplication factor so that there is a balanced distribution of examples from each of them. For English, we do not balance the number of examples between the reconstructed BooksCorpus and Wikipedia, as no comparable balancing was applied in the pre-training of the original BERT. Similar to \cite{virtanen2019multilingual}, whole-word masking is used, and other parameters and process for data creation match those in \cite{devlin2018bert}.

\subsection{Pre-training process}

We primarily follow the pre-training process and implementation\footnote{\url{https://github.com/TurkuNLP/FinBERT/blob/master/nlpl_tutorial}} described in \cite{virtanen2019multilingual}. Briefly, the model architecture is that of \bertbase{}, with 110M parameters excluding the word embeddings.
The model is trained for 1M steps. For the first 0.9M steps of training, we use a sequence length of 128 and batch size of 140. For the remaining 0.1M steps, the sequence length is set to 512 and (by contrast to FinBERT training), a batch size of 16 is used due to memory constraints.
We use the LAMB optimizer \citep{You2020LAMB} with warmup over 10K steps to a learning rate of 1e-4 followed by decay. The model is trained on 8 Nvidia V100 GPUs for approximately 12 days.

\section{Evaluation}
We evaluate \bbert{} on the English and Finnish benchmarks that have been used to evaluate the corresponding monolingual BERT models to allow direct comparison of the performance of the bilingual and monolingual models. For English, we choose the GLUE benchmark. 
For Finnish, we choose the benchmarks that have been used to evaluate \finbert{}. These include the following tasks: POS tagging, NER, dependency parsing, and text classification tasks. We follow the procedures used in \cite{virtanen2019multilingual} for Finnish evaluation.

\subsection{English}

\subsubsection{Data}
The General Language Understanding Evaluation (GLUE) benchmark \citep{wang2018glue} is a collection of nine NLU datasets. The test sets of some of these data sets are only available through the online GLUE evaluation server\footnote{\url{https://gluebenchmark.com/}}. Following \cite{devlin2018bert}, we exclude the Winograd Schema Challenge dataset \citep{levesque2011winograd} from evaluation, using the following eight datasets:
\begin{itemize}
  \item \textbf{CoLA} The Corpus of Linguistic Acceptability \citep{warstadt2018neural} is a collection of sentences from published linguistics literature. Its training set consists of 8.5K examples, each a sentence with a binary label indicating whether the sentence is linguistically acceptable or not.
  \item \textbf{SST-2} The Stanford Sentiment Treebank \citep{socher2013recursive} consists of a collection of sentence excerpts from movie reviews, with a training set size of 67K examples. The GLUE benchmark uses sentence-level human judgments of sentiment, i.e., each example consists of a sentence with a label, either positive or negative.
  \item \textbf{MRPC} The Microsoft Research Paraphrase Corpus \citep{dolan2005automatically} consists of sentence pairs from online news. Its training set has 3.7K examples, each a pair of automatically extracted sentences and a human judgment of whether they are semantically equivalent.
  \item \textbf{QQP} The Quora Question Pairs\footnote{\url{https://www.quora.com/q/quoradata/First-Quora-Dataset-Release-Question-Pairs}} is a collection of question pairs extracted from the Quora website. The training set has 364K examples, each a pair of questions and judgment of whether they are semantically equivalent.
  \item \textbf{STS-B} The Semantic Textual Similarity Benchmark \citep{Cer2017STS} consists of sentence pairs taken from various sources such as news headlines and image captions. The training set has 7K examples, each a pair of sentences and a human judgment of their similarity, ranging from 1 to 5.
  \item \textbf{MNLI} The Multi-Genre Natural Language Inference Corpus \citep{williams2018broad} is a collection of crowd-sourced sentence pairs and human judgments of their textual entailment relations (\texttt{entailment}, \texttt{contradiction}, or \texttt{neutral}). The corpus has a training set of 383K examples, and in- and out-of-domain development and test sets. The in-domain data (matched) are drawn from the same genres as the training data, while out-of-domain data (mismatched) are drawn from different genres.
  \item \textbf{QNLI} The Stanford Question Answering Dataset \citep{rajpurkar2016squad} is a set of question and answer pairs converted into the NLI format. The questions are collected from Wikipedia, while the answers are written by annotators. The training set of QNLI consists of 105K examples, each a question, an answer, and a label (either \texttt{entailment} or \texttt{not\_entailment}).
  \item \textbf{RTE} The Recognizing Textual Entailment (RTE) datasets are compiled from the datasets of four textual entailment challenges: RTE1 \citep{dagan2006pascal}, RTE2 \citep{bar2006second}, RTE3 \citep{giampiccolo2007third}, and RTE5 \citep{bentivogli2009fifth}. The training set consists of 2.5K examples drawn from news and Wikipedia, and each example has two sentences and a label (\texttt{entailment} or \texttt{not\_entailment}).
\end{itemize}

\subsubsection{Methods}
We follow the fine-tuning approach introduced by \cite{devlin2018bert} for hyperparameter selection and additional task-specific additions to the model architecture\footnote{The implementation published at \url{https://github.com/google-research/bert} is used.}. For hyperparameter selection, we use a batch size of 32 and fine-tune for 3 epochs for all GLUE tasks. For each task, we search for the best performing learning rate among \{2e-5, 3e-5, 4e-5, 5e-5\} on the development set with three replicates. For all tasks, a task-specific layer is added after the final Transformer layer. For tasks with two input sentences, both sentences are given as inputs at a time, in the form of \texttt{[CLS] Sentence A [SEP] Sentence B}, as in pre-training. For tasks with one input sentence, the second sentence is seen as degenerate, and the input is only \texttt{[CLS] Sentence A}.

\subsubsection{Results}

The results on the test sets are shown in Table~\ref{tab:glue}. Overall, \bbert{} performs comparably with Google's BERT. While most of the differences are within half a percentage point, the original BERT obtains better results with single sentence classification tasks, and \bbert{} performs better on the RTE task.

\begin{table}[!t]
  \begin{tabular}{llllllllll}
    \toprule
    System & CoLA & SST-2 & MRPC & QQP & STS-B & MNLI-(m/mm) & QNLI & RTE & \textbf{Average} \\
    & 8.5K & 67K & 3.5K & 363K & 5.7K & 392K & 108K & 2.5K & - \\
    \midrule
    \bertbase{} & \textbf{52.1} & \textbf{93.5} & \textbf{88.9} & 71.2 & 85.8 & \textbf{84.6}/83.4 & 90.5 & 66.4 & 79.6 \\
    \texttt{B-BERT} & 48.7 & 92.9 & 88.4 & \textbf{71.6} & \textbf{86.1} & 84.4/\textbf{83.8} & \textbf{91.0} & \textbf{69.5} & 79.6 \\
    \bottomrule
  \end{tabular}
  \caption{Results on GLUE test sets scored by the evaluation server (\url{https://gluebenchmark.com/leaderboard}). The row of numbers under the task name is the size of the training set. For CoLA, the metric is Matthew's correlation, and F-score for MRPC and QQP, Spearman correlation for STS-B, and accuracy for the rest of the task. The average scores are higher than those reported on the GLUE leaderboard because the WNLI task is excluded. The \bertbase{} results are taken from \cite{devlin2018bert}.}
  \label{tab:glue}
\end{table}

\subsection{Finnish}
We use four tasks that have been used in \finbert{} evaluation to evaluate \bbert{}: POS tagging, NER, dependency parsing, and text classification.

\subsubsection{Data}
\label{sec:Finnish-data}

We briefly introduce the datasets used 
for these four tasks in this section and refer to \cite{virtanen2019multilingual} for detailed descriptions of the data.

\begin{itemize}
  \item \textbf{POS tagging} The three Finnish treebanks in the Universal Dependencies (UD) collection \citep{Nivre2016UD} are used: the Turku Dependency Treebank (TDT) \citep{Pyysalo2015FinUD}, FinnTreeBank (FTB) \citep{Voutilainen2012FTB}, and the Parallel Universal Dependencies treebank (PUD) \citep{Zeman2017conll}. The UD version 2.2 is used for comparability with the \finbert{} results, which in turn chose this version for comparability with the results of the CoNLL shared tasks in 2017 and 2018 \citep{Zeman2017conll,Zeman2018conll}. For the PUD corpus, which has no training or development sets, the corresponding sets from the TDT corpus are used for training and parameter selection.
  \item \textbf{NER} \finbert{} was only evaluated on the FiNER \citep{Ruokolainen2020FiNER} corpus because it was the only NER corpus available for Finnish during its development. FiNER has its text source from a Finnish online technology news, and contains CoNLL-style annotations of person, organization, location, product, and event, as well as dates. It also contains an out-of-domain test set, where the text source is Wikipedia. Though the corpus contains a small number of nested annotations, we follow the \finbert{} evaluation and use only non-nested annotations.
  \item \textbf{Dependency parsing} The same versions of the three Finnish UD treebanks used for POS-tagging are used also for this task. 
  \item \textbf{Text classification} We use the text classification corpora created by \cite{virtanen2019multilingual}. These are the Yle news corpus, which contains formal text, and the Ylilauta corpus, whose text source is the online forum Ylilauta, which thus contains informal text. The Yle news corpus contains documents, each annotated with one of ten topics, such as sports, politics, and economy. Due to license restrictions, the corpus cannot be redistributed, but the code for recreating the dataset is available.\footnote{\url{https://github.com/spyysalo/yle-corpus}} The Ylilauta corpus contains online postings and the topic of the boards onto which they were posted. As with the Yle news corpus, the Ylilauta corpus also contains documents for ten topics. Both corpora have training set size of 100K and balanced distribution of classes. All of the documents are truncated to at most 256 tokens to reduce the advantage that a compact representation for Finnish may introduce for language-specific models. 
\end{itemize}

For ease of comparison of model performance, we organize the results similarly to the GLUE benchmark to obtain a single number for each model. Specifically, there are four tasks, and each task has its own subsets: POS tagging (TDT, FTB, and PUD treebanks), NER (in-domain and out-of-domain test sets), dependency parsing (TDT, FTB, and PUD treebanks), and text classification (Yle and Ylilauta corpora). For dependency parsing, \cite{virtanen2019multilingual} report results for both predicted and gold segmentation, whereas we choose the results for gold segmentation to focus specifically on model performance for parsing. For text classification, \cite{virtanen2019multilingual} also evaluate on downsampled versions of the corpora, but we include only the full versions of the corpora. We adopt a similar approach to GLUE and take the algebraic mean as the final score for comparison, though we recognize that (as in GLUE) these evaluations use different metrics.

\subsubsection{Methods}
We follow the task-specific model architectures implemented in \cite{virtanen2019multilingual} for all the tasks.

\begin{itemize}
    \item \textbf{POS tagging} The BERT POS tagger\footnote{\url{https://github.com/spyysalo/bert-pos}} adds a time-distributed dense output layer on top of the BERT model and represents each word by its first WordPiece-tokenized input word. The official CoNLL 2018 evaluation script is used, which reports the UPOS metric.
    \item \textbf{NER} The NER implementation\footnote{\url{https://github.com/jouniluoma/keras-bert-ner}} follows the method employed by \cite{devlin2018bert}, which attaches a dense layer over BERT and independently predicts IOB tags. The evaluation metric is mention-level F-score, as implemented in the standard \texttt{conlleval} script. 
    \item \textbf{Dependency parsing} UDify \citep{kondratyuk2019Udify} is a multi-task model for BERT that jointly predicts POS tags, morphological features, lemmas, and dependency trees. It has task-specific prediction layers on top of BERT, as well as task-specific layer attention that calculates the weighted sum of representations for each token from all layers. The original UDify model is fine-tuned on \mbert{} and was trained on 75 languages with UD treebanks. We use \bbert{} and train separately on the TDT and FTB treebanks. As PUD has no training set, the model trained on TDT is used for prediction on the PUD test set.  The metric used for parsing is Labeled Attachment Score (LAS). 
    \item \textbf{Text classification} A task-specific layer on top of the \texttt{[CLS]} token is added for class prediction following \cite{devlin2018bert} and \cite{virtanen2019multilingual} and performance is evaluated in terms of accuracy.
\end{itemize}

Unless otherwise specified, parameter selection is carried out the same way as \cite{virtanen2019multilingual}, with a grid over the learning rate, the number of epochs, and the batch size. For parameter selection on the development sets, 3-5 replicates are run, and 5-10 replicates are run with the selected parameters for final evaluation on the test sets. No replicates are done for dependency parsing due to the large training set. For text classification, the grid is narrowed down to batch size of 16 and 20, while fixing the learning rate at 2e-5 and number of epochs at 4 based on previous experiments.

\subsubsection{Results}
The results are reported in Table~\ref{tab:finnish-results}. Overall, \finbert{} achieves the best average performance of 92.34\%, while \bbert{} scores slightly below at 92.02\%. Although the performance does not match that of \finbert{}, \bbert{} performs almost as well compared to the overall score of \mbert{} at 88.88\%; \bbert{} covers 90.75\% of the performance difference between \mbert{} and \finbert{}.
The slightly lower scores obtained by \bbert{} could be due to randomness in training, or the fact that the effective number of tokens for Finnish is smaller for \bbert{} than \finbert{}, as the 80,000 tokens are shared between English and Finnish.

We note that although results for POS tagging, NER, and dependency parsing results were also recently reported by \cite{Ulcar2020FinEstBERT} for their Finnish-Estonian-English BERT model, these results are not directly comparable to ours: the UD version of the treebanks they used were not specified for POS tagging and dependency parsing, and although they used the same corpus (FiNER) for NER, the named entity types were reduced to only person, location, and organization. We thus refrain from direct numerical comparison to their results.

\begin{table}[!t]
  \centering
  \begin{tabular}{llllll}
    \toprule
    & POS tagging & NER & Dependency parsing & Text classification & Average\\
    & (TDT/FTB/PUD) & (ID/OOD) &  (TDT/FTB/PUD) & (Yle/Ylilauta) &  \\
    \midrule
    \finbert{} & \textbf{98.23}/\textbf{98.39}/\textbf{98.08} & \textbf{92.40}/\textbf{81.47} & \textbf{93.56}/\textbf{93.95}/\textbf{93.10} & \textbf{91.74}/\textbf{82.51} & \textbf{92.34} \\
    \bbert{} & 98.14/98.16/98.07 & 92.23/81.08 & 93.16/93.50/93.02 & 91.37/81.42 & 92.02 \\
    \mbert{} & 96.97/95.87/97.58 & 90.29/76.15 & 87.99/87.46/89.75 & 90.28/76.51 & 88.88 \\
    \bottomrule
  \end{tabular}
  \caption{Results for evaluation on all Finnish tasks. All the BERT models are cased models. The metric for POS tagging is UPOS, NER F-score, dependency parsing LAS, and text classification accuracy. Apart from the \bbert{} numbers, the other results are taken from \cite{virtanen2019multilingual}. For NER, ID stands for the in-domain test set (news), and OOD stands for the out-of-domain test set (Wikipedia).}
  \label{tab:finnish-results}
\end{table}

\section{Discussion and future work}
\label{sec:discussion}
Multilingual BERT models have been shown to learn a language-agnostic subspace that allows cross-lingual transfer \citep{Wu2019xlingBERT,Pires2019HowMBERT,Cao2020mBERTAlignment}. However, current research has shown that model capacity needs to be increased to cover multiple languages without loss in model quality \citep{Conneau2019XLMR}. There are several potential ways in which BERT capacity could be increased. Taking into account the observation by \cite{Artetxe2020cross} that the effective vocabulary size per language plays an important role in model performance, we have here explored an approach where the vocabulary size was increased without increasing the model size in other ways or increasing the number of training steps.

Setting the vocabulary size to be the sum of the sizes of those of the monolingual models, we trained a Finnish-English bilingual BERT model and compared its performance to English and Finnish monolingual BERTs on various established benchmark tasks.
We have shown that it is possible to create a bilingual model without compromising the performance on either language.
As multilingual BERT models have been shown to be capable of cross-lingual transfer \citep{Artetxe2020cross,Ulcar2020FinEstBERT}, we expect that our bilingual model can be used for cross-lingual tasks as well as for the monolingual tasks considered here, but direct study of the cross-lingual capabilities is currently hindered by the lack of Finnish-English cross-lingual datasets.
Our approach increased vocabulary size but not other aspects of the model, leaving other potential approaches as future work.
We note that, during training, the bilingual BERT saw roughly half of the number of examples for Finnish as did the Finnish BERT, since the total number of examples seen by both models are approximately the same. The ability of bilingual BERT to perform almost on par with the Finnish BERT may be due to the language-agnostic subspace reported to be learned by BERT models.

Potential follow-up questions to our study include: to what degree the approach of increasing vocabulary size can be extended to cover more languages, and whether the effective vocabulary size per language to achieve comparable performance remains the same when the number of languages increases, if comparable performance is possible.
One limitation of our evaluation is the use of the GLUE benchmark without evaluation on more challenging datasets or benchmarks such as SQuAD or SuperGLUE \citep{wang2019superglue}. Question answering is considered to be a more challenging task, whereas most of the tasks in GLUE are two- or three-way classification tasks, some of which have been criticized for the presence of artifacts \citep{Gururangan2018artifacts}.
We hope to explore these questions as well as the cross-lingual capabilities of the model in future work.




\section{Conclusions}

We have studied the feasibility of training a fully bilingual deep neural language model, i.e.\ a model that approaches or matches the performance of monolingual models at language-specific tasks. We trained a bilingual Finnish-English \bertbase{} model by expanding the vocabulary size to be the sum of the size of the two individual vocabularies, and compared the model performance to monolingual models. We found that, on a range of NLU tasks, the bilingual model performs comparably or nearly comparably with monolingual models. We conclude that, for the \bertbase{} architecture, it is possible to train a fully bilingual deep contextual model for two remotely related languages. We release the newly introduced \bbert{} model and all tools introduced to create the model under open licenses at \url{https://github.com/TurkuNLP/biBERT}.

\section*{Acknowledgements}
We are thankful for the support of CSC IT Center for Science for the computing resources it provides. L.H.C. is founded by the DigiCampus project overseen by the EXAM Consortium.
We are grateful to Juhani Luotolahti and Antti Virtanen for the technical help they provided. We also thank Jouni Luoma for his consultation on the Finnish NER experiments.

\bibliography{ms}

\end{document}